\RequirePackage[hyphens]{url}

\documentclass[10pt]{article} 
\usepackage[preprint]{tmlr}

\usepackage{amsmath,amsfonts,bm}

\def\eqref#1{equation~\ref{#1}}

\def\1{\bm{1}}

\DeclareMathAlphabet{\mathsfit}{\encodingdefault}{\sfdefault}{m}{sl}
\SetMathAlphabet{\mathsfit}{bold}{\encodingdefault}{\sfdefault}{bx}{n}

\usepackage{hyperref}
\usepackage{url}
\usepackage{amsmath}
\usepackage{dsfont}
\usepackage{booktabs}
\usepackage{multirow}
\usepackage{float}
\usepackage{graphicx}
\usepackage{makecell}
\usepackage{placeins}
\usepackage{mdframed}
\usepackage{listings}
\title{A Formal Framework for Uncertainty Analysis of Text Generation with Large Language Models}

\author{%
\name Steffen Herbold$^*$ \email steffen.herbold@uni-passau.de \\
\addr Faculty of Computer Science and Mathematics\\
University of Passau\\
Passau, Germany
\AND
\name Florian Lemmerich$^*$ \email florian.lemmerich@uni-passau.de \\
\addr Faculty of Computer Science and Mathematics\\
University of Passau\\
Passau, Germany
}

\def\month{03}  
\def\year{2026} 
\def\openreview{\url{https://openreview.net/forum?id=TODO}} 

\begin{document}

\maketitle
\centerline{\footnotesize $^{*}$ Authors are equal contributors and listed alphabetically.}

\begin{abstract}
The generation of texts using Large Language Models (LLMs) is inherently uncertain, with sources of uncertainty being not only the generation of texts, but also the prompt used and the downstream interpretation. Within this work, we provide a formal framework for the measurement of uncertainty that takes these different aspects into account. Our framework models prompting, generation, and interpretation as interconnected autoregressive processes that can be combined into a single sampling tree. We introduce filters and objective functions to describe how different aspects of uncertainty can be expressed over the sampling tree and demonstrate how to express existing approaches towards uncertainty through these functions. With our framework we show not only how different methods are formally related and can be reduced to a common core, but also point out additional aspects of uncertainty that have not yet been studied. 
\end{abstract}

\section{Introduction}

Large Language Models (LLMs) are modeling language through a large, complex probability distribution such that the generation of texts can be interpreted as stochastic processes. This inherent uncertainty allows us to sample different results, which enables a certain desirable diversity in outcomes~\citep{sandholm2024randomness, wang2025effect}, but is also directly related to the existence of hallucinations as a property of these processes~\citep{kalai2024calibrated, xu2024hallucination, kalai2025language}. Therefore, the uncertainty associated with information generated by an LLM is a key property to understand an LLM's capabilities. However, the uncertainty in the LLM-based text generation process is not limited to the generation process: the wording of the used prompt and the interpretation of the generated outcome are also sources of uncertainty~\citep{xia2025survey}.

While there are many works looking at different aspects of uncertainty (see Section~\ref{sec:related-work}), there is no work yet that formally aligns different approaches that consider uncertainty from a general perspective that includes the prompt, the generated texts, and the interpretation of the generated text. Within this work, we address this gap in the literature through the development of a formal framework that allows us to describe different variants of probability or entropy-based uncertainty definitions and measurements. 

Our framework is based on the notion of a \textit{sampling tree}, i.e., the assumption that the prompt, the generated text, and the interpretation are all generated auto-regressively within a single stochastic process. 
We introduce the notions of \emph{filters} to define which parts of the sampling tree are used to assess the uncertainty. We also define \emph{objectives} to generalize the probability of the sampling to be able to account for semantics. We proceed to demonstrate how different approaches from the state-of-the-art can be expressed in our framework and show how they relate to each other. Afterwards, we explore aspects of uncertainty that our framework models, but that are not yet within the focus of the state-of-the-art to provide guidance for future work. Through this, our work contributes not only a formal foundation to understand the relationship between approaches, but also an understanding of blind-spots within the current literature:
\begin{itemize}
    \item How the uncertainty induced by prompts and generated texts further propagates into the uncertainty of judgments of the outputs by actors (agentic or human) is so far not considered, but could be addressed using the same techniques. 
    \item Different approaches for introducing the semantics into uncertainty quantification are closely related to each other and can be interpreted as hard-clustering and soft-clustering methods used to rescale the probabilities similar to the expectation maximization algorithm. 
    \item There is only a limited understanding of how sampling -- both during the generation, as well as for the estimation of uncertainty -- affects the uncertainty estimates, including how sampling during generation and during estimation are related to each other. 
    \item The formal relationships between different uncertainty quantification approaches that our framework identifies are not sufficiently considered by current taxonomies, highlighting the need for a systematic taxonomy that enables both a clear definition of approach families (e.g., syntactic, semantic, probability-based or entropy-based) as well as the scope of uncertainty that is considered by each approach (e.g., whether the impact of the prompt on the output, the output format, or the interpretation is considered).
\end{itemize}
Overall, our work enables a unified conceptualization of LLM uncertainty that bridges previously isolated approaches under a single formalized framework and provides a clear roadmap to address critical blind spots in future research.

\section{Related Work}
\label{sec:related-work}

Measuring uncertainty of LLM outputs has become an increasingly relevant and popular research topic recently as limited uncertainty is seen as a pre-requisite for reliable and safe models, e.g., \cite{malinin2020uncertainty, kuhn2023semantic, zhang2023enhancing, huang2025look}. Often, uncertainty in processes can be categorized into \emph{aleatoric} uncertainty, i.e., irreducible uncertainty due to the nature of the statistical process, and \emph{epistemic} uncertainty \citep{hullermeier2021aleatoric} due to a lack of knowledge. With LLMs, epistemic uncertainty could be due to lack of knowledge about something during training, unclear goals when formulating a prompt, or different backgrounds of judges interpreting outcomes. Aleatoric uncertainty could be due to unconscious decisions by human actors or the decoding by LLMs. 
However, this distinction is not always clear, and for many practical applications irrelevant. Moreover, when publications distinguish between aleatoric and epistemic uncertainty, the general method to compute the underlying probabilities/entropies is the same for both~\citep[e.g.,][]{aichberger2024rethinkinguncertaintyestimationnatural}.
Thus, for this paper we drop this distinction in favor of a summarized formalism to quantify uncertainty.

Due to the exploding number of proposals to measure uncertainty,  some recent papers aimed to summarize and categorize these approaches:
\cite{yang2025understanding} provide a framework to decompose the total uncertainty into different aspects including prompting, context, and preprocessing of multi-modal inputs. In contrast to our approach, they do not consider different types of estimators that can be used to measure the uncertainty of each component but focus on a single entropy-based measure. \cite{liu2025uncertainty} present a general survey on uncertainty quantification and confidence calibration in the context of LLMs. They distinguish between different sources of uncertainty: (i) uncertainty induced by the input (not uncertainty of the input itself), (ii) uncertainty of reasoning steps, (iii) uncertainty of the trained model parameters, and (iv) prediction uncertainty stemming from different sampling runs. Our framework shares some described concepts but focuses on the uncertainty measurement for one given models and prompt intentions and introduces a formal framework that integrates different measurement options.

\cite{xia2025survey} summarize different avenues for estimating uncertainty in LLMs in form of a taxonomy including latent information, consistency-based, and semantic clustering methods. While similar approaches are discussed together, the taxonomy does not establish a common formal foundation for different methods. Despite using a somewhat different categorization of concrete measures, our approach can mostly cover these approaches in a unifying formal framework.

\section{Sources of Uncertainty}
\label{sec:sources-of-uncertainty}

\begin{figure}[t]
    \centering
    \includegraphics[width=1.0\linewidth]{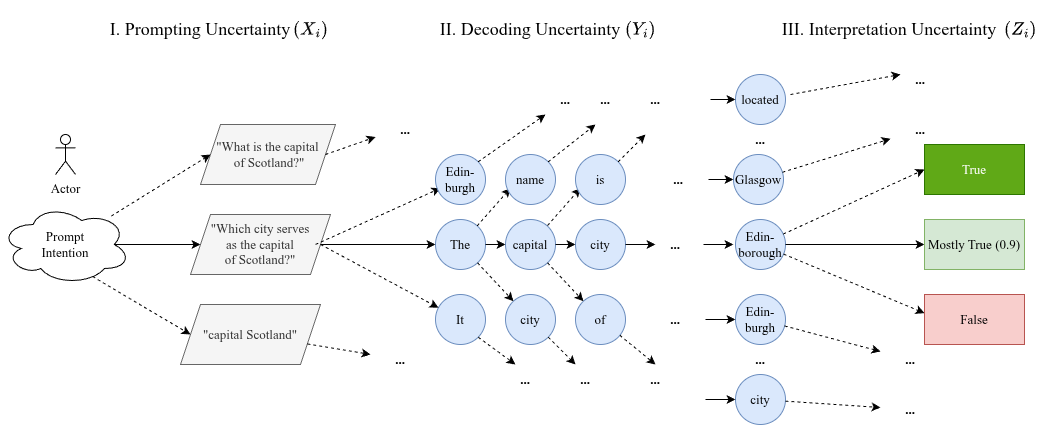}
    \caption{Structured view of generating an LLM response with corresponding uncertainties. In our formal framework, the random variables $X_i, Y_i,$ and $Z_i$ are corresponding to modeling the uncertainty of the different phases of the generative process.}
    \label{fig:pipeline}
\end{figure}

As discussed above, the different approaches from the literature to quantify uncertainty consider many aspects with respect to the generated output, including the prompt, the decoding, and judgments of generated content. However, they focus on specific aspects of the generation process such as the uncertainty originating from the sampling process in the decoding or the sensitivity to small variations in the prompt. In our paper, we aim to unify these measures by considering the overall process of an (agentic or human) actor getting a response from an LLM as a multi-step process, cf. Figure~\ref{fig:pipeline}. Within this section, we describe our conceptual view on uncertainty to describe which aspects need to be considered, before developing this into a formal framework in Section~\ref{sec:formalism}.

\subsection{Uncertainty in the generative process}

The conceptual foundation of our work is a structured view on the main generation process with LLMs that has three main steps: (i) the generation of the concrete prompt from an actor's intention; (ii) the generation of the LLM response through sequential decoding; and (iii) the interpretation of this response by generating a judgment. Each step corresponds to respective sources of uncertainty, which we will develop into the concept of a \textit{sampling tree} in our formal framework.

\subsubsection{Prompting uncertainty}

Actors typically employ LLMs with a specific intent (e.g., ``finding the capital of Scotland'') and expect the LLM to generate solutions for their specified tasks.
However, the concrete prompt that is given to the LLM to complete this task varies depending on the user and technical context of the model.
In particular, the specific prompt of a human actor might depend on conscious as well as unconscious choices such as (i) language choice by a multilingual user; (ii) exact prompt phrasing of the question; (iii) prompt formatting; (iv) typographic mistakes; and (v) the general user-context, such as general instructions from the user-provided system-prompt. Similarly, agentic actors have a context that influences the prompt generation. 
From an analysis point of view, the actor's behavior is inherently non-deterministic but subject to uncertainty through unknown characteristics, spontaneous decisions, or human errors.
For example, regarding the capital of Scotland, the prompt could be ``Which city serves as the capital of Scotland?'', ``What is the capital of Scotland?'' or just ``capital scotland'' without consciously deliberating over the phrasing or due to differences in how an agentic approach generates the prompt.


\subsubsection{Decoding uncertainty}

In current generative LLMs, the core inference step, the token-by-token generation of text, involves inherent randomness in the decoding. Typically, the next token is selected by sampling from a probability distribution of candidates, e.g., directly from the probability distribution of the next token, by nucleus sampling~\citep{holtzman2020curious} or top-$k$ sampling \citep{fan-etal-2018-hierarchical}. Prior to sampling, the temperature parameter enables the re-scaling of the output probabilities to increase randomness (high temperature) or the likelihood of high-probability tokens (low temperature). From an uncertainty perspective, the choice of temperature effectively means that a different language model is used. 

For example, the LLM might generate the output ``The capital of Scotland is Edinburgh.'' instead of ``It is Edinburgh.'' or maybe even a completely different result like ``Glasgow'' just due to random decisions in the decoding algorithm.

\subsubsection{Interpretation uncertainty}

LLM results are typically consumed by a downstream actor, which interprets the generated result. This actor can be the same as the one responsible for the prompt generation or different. A human actor could be an annotator who assigns a code to the data or describes the outcome, but also a reader who reads the response and then creates their own interpretation. For simplicity, we do not differentiate between whether human actors write down, verbalize or simply form an interpretation through thoughts in their head. We only assume that this is somehow put into words. An agentic actor could be an LLM-as-a-judge approach, who takes the outcome of the decoding as input (together with additional configuration like prompts guiding the judgment) and generates the interpretation. 


Both options will come with their respective uncertainties, either through human subjectivity and biases or by the uncertainty of the LLM used as a judge. On top of that, the choice of annotator/interpreter itself can be seen as a random process as it depends often on practical considerations independent of actual analysis task.

\subsection{Focus of uncertainty analysis}

The structured view we introduced above allows us to capture uncertainty across the whole process. In practice, we need to focus the uncertainty analysis to specific aspects of this process. This is also the case in the literature on uncertainty, which, e.g., focuses on the prompt, the decoding, or the interpretation. Whenever the uncertainty of LLM responses is analyzed, it must be clearly decided and stated, which components in this random process are fixed and which components are varied (i.e., sampled from a random process). For example, one can analyze the prompt uncertainty with a fixed (seeded) decoding without considering the interpretation or one can analyze the combined prompting and decoding uncertainty by varying both the prompt phrasing and the decoding samples while keeping the interpreter fixed. Obviously, there are many more examples for the focus. The formal framework we develop below introduces \textit{filter} functions to enable such a clear definition. 

\subsection{Syntax, semantics, and other considerations}

When generative LLMs are analyzed in terms of uncertainty, typically this is done with respect to a particular aspect, e.g., by considering the exact wording through the syntax, analyzing aspects of the semantic similarity, or by seeing if a certain expected structure is used. We call this an \emph{objective} of the uncertainty analysis.
That is, we investigate with respect to \emph{which} property the LLM actually generates a robust response. For example, we could focus on the syntax by considering only the exact identity of responses as equivalent and, therefore, measure the uncertainty with respect to the exact model response. However, we could also be interested in outcomes with the same semantics. In this case we would consider, e.g., different spelling (``Edinburgh'' vs ``Edinborough'') of the same entity as equivalent, and therefore the model as certain if it always returns either of those outcomes.

This concept is very broad and allows the analysis of the uncertainty to take many different types of objectives into account such as consistency of the used tone or style, the consistency of the applied logical steps, referring to the same main facts, or generally just measuring if the provided answer is correct or not.
In order to ensure a precise notion of uncertainty, an analyst needs to specify this objective, similar to the need of defining the focus. To facilitate this, we develop the mathematical notation of \textit{objective functions} within our formal framework.

\section{A Formal Framework for Uncertainty in LLM Inference}
\label{sec:formalism}

Within this section, we present our core contribution: a formalism that allows us to provide a generalized definition of the probability and entropy of outcomes from an LLM sampling process, including the interpretation by a judge of the output. This formalism is based on the concept of a sampling tree that models the probabilities associated with all possible output sequences from an LLM. Objectives and filters specify how this sampling tree is used to compute probabilities and the entropy of outcomes. We will later show in Section~\ref{sec:operationalization} how these concepts serve as core building blocks for concrete uncertainty measurements. 

\subsection{Basic Notations}
\label{sec:basic-notations}

To define a sampling tree, we model both the prompt, generated text, and the interpretation as random variables. Let $V = \{1, \dots, b\}$ be the vocabulary of tokens, where $b \in \mathbb{N}$ is the vocabulary size. Tokens are natural numbers without loss of generality, though we use word representations in examples below. Let $\mathcal{V}$ denote the set of all finite token sequences.

Let $X_i, Y_j, Z_k$ be discrete random variables over $V$ with $i \in \{1,\dots, n_x\}$, $j \in \{1,\dots,n_y\}$, $k \in \{1, ..., n_z\}$,  and $n_x, n_y, n_z \in \mathbb{N}$. Within this paper, we assume that the $X_i$ represent the prompt, $Y_j$ represent the generated output, and $Z_k$ the interpretation by a judge. Thus, the random variables result in a sequence $(X_1,\dots, X_{n_x}, Y_1,\dots, Y_{n_y}, Z_1, \dots, Z_{n_z})$. For brevity, we use the notations $\mathbf{X}=(X_1, \dots, X_{n_x})$, $\mathbf{Y}=(Y_1, \dots, Y_{n_y})$, $\mathbf{Z}=(Z_1, \dots, Z_{n_z})$ for the complete prompt, output, and interpretation. In cases where we do not want to distinguish between the prompt, output, and interpretation, we use $\mathbf{V}=(V_1, ..., V_n)$ with $n\in \mathbb{N}$ to identify a sequence of tokens as random variables. 

The notation $\mathbf{V}_{p:q}=(V_p, \dots, V_q)$ represents a subsequence from index $p$ to $q$. 
We use the lower case letters $\mathbf{x}$, $\mathbf{y}$, $\mathbf{z}$, and $\mathbf{v}$ for sampled tokens with the same index notations as for the random variables. E.g., $\mathbf{x}=(x_1, \dots, x_{n_x})$ represents a concrete prompt and $\mathbf{v}=(v_1, \dots, v_n)$ represents a sequence of tokens without a specific interpretation. Further, we use $|\mathbf{v}|$ as notation for the length of a sequence.

Based on these notations and the autoregressive nature of LLM-based text generation with decoder-only models, the probability of the output token $Y_i$ is
\begin{equation}
P(Y_i|\mathbf{X},\mathbf{Y}_{1:i-1})    
\end{equation}
in general and
\begin{equation}
p(y_i|\mathbf{x},\mathbf{y}_{1:i-1})= P(Y_i=y_i|\mathbf{X}=\mathbf{x},\mathbf{Y}_{1:i-1}=\mathbf{y}_{1:i-1})    
\end{equation}

for a concrete prompt $\mathbf{x}$, and already generated sequences of the output $\mathbf{y}_{1:i-1}$. For the complete output $\mathbf{y}$ we write $p(\mathbf{y}|\mathbf{x})$. Similarly, the interpretation depends on the output and we write $p(\mathbf{z}|\mathbf{x},\mathbf{y})$. Moreover, we assume that the relationship between $X_i$ and $Z_k$ is also auto-regressive, i.e., that prompts and interpretations are created left-to-right as well. Based on our interpretation of the process introduced in Section~\ref{sec:sources-of-uncertainty}, this is a natural assumption for LLMs as actors. For humans, this might not always be the case, as they may modify texts in between. We discuss how this could be resolved in Section~\ref{sec:measurement-uncertainty}.

\subsection{Sampling tree}
We can visualize the overall process of generating the output of an LLM as a sequential sampling process as a \emph{sampling tree} $\mathcal{T}$. 
In this tree, the root node represents an empty output sequence. Each tree node $v$ at depth $i$ corresponds to a concrete partial output $\mathbf{v}_{1:i} \in \mathcal{V}$.

The construction of our sampling tree happens in three stages:
\begin{enumerate}
    \item the process of sampling the exact prompt $\mathbf{X}$ based on the original intent. The uncertainty in this stage originates from indeliberate human choices or variability of an employed model or method for generating prompt alternatives.
    \item the sampling process to generate the model output tokens $\mathbf{Y}$ conditioned on the chosen input prompts $\mathbf{X}$.
    \item the process of producing an interpretation $\mathbf{Z}$ of the output given an input prompt $\mathbf{X}$ and a model output $\mathbf{Y}$.
\end{enumerate}

However, within the sampling tree we do not need to distinguish between $\mathbf{X}, \mathbf{Y}, \mathbf{Z}$ and as these are all just token sequences with a fixed order ($\mathbf{X}, EOS_X, \mathbf{Y}, EOS_Y, \mathbf{Z}, EOS_Z$). Here, $EOS_X$ represents the end of the prompt, the $EOS_Y$ represents the end-of-sequence token of the output and $EOS_Z$ the end-of-sequence token of the interpretation. These tokens only serve the role of separating different parts of the sequence and they are not the same as the EOS sentinel token, that is often included in LLM vocabularies to indicate that an LLM can stop generating output. Instead of naming these concepts explicitly, we can just work with the notation of a general sequence of tokens $\mathbf{V}$ that we introduced above when discussing aspects regarding the sampling tree. 

Within the sampling tree, each edge is annotated with the probability to extend the direct prefix to the sequence represented by the target node, e.g., the edge between $\mathbf{v}_{1:i-1}$ and $\mathbf{v_i}$ is labeled with $p(v_i|\mathbf{v}_{1:i-1})$.
Since each node (except the root) has exactly one incoming edge, we can define for more concise notation for each
\begin{equation}
\label{eq:path-prob}
pr_{\mathbf{v}_{1:i}} = p(v_i|\mathbf{v}_{1:i-1})    
\end{equation}
as the probability of the incoming edge into the node $\mathbf{v}_{1:i}$.

To deal with sequences of different lengths, we define $P(V_i=EOS_Z|V_{i-1}=EOS_Z)=1$ 
and $P(V_i\neq EOS|V_{i-1}=EOS_Z)=0$, i.e., the probability of tokens other than $EOS_Z$ is zero once an $EOS_Z$ appeared in a sequence of tokens.

\subsection{Filters for the sampling tree}

Often, we do not want to work with the full sampling tree, but rather with a subtree, e.g., specifying fixed prompt, focusing only the output, without considering the interpretation. To enable this, we define a filter function to specify to which token sequences this objective is applied as
\begin{equation}
    f: \mathcal{V} \to \mathbb{N}_0
\end{equation}
mapping to the natural numbers including zero. We can further define the subset of all filtered sequences $\mathcal{V}^{f}= \{\mathbf{v} \in \mathcal{V}: f(\mathbf{v})>0\}$. 

Note, that in most cases, filters will be binary, i.e., mapping to $[0,1]$ to select a specific part of the sampling tree - which is also the reason why we refer to this concept as filter. However, there are also cases where the same path in the sampling tree is considered multiple times during the uncertainty estimation (see Section~\ref{sec:monte-carlo}), which we enable by instead defining the map as a mapping to the natural numbers. This can be interpreted as the number of times a specific path $\mathbf{v}\in \mathcal{V}$ is passing through the filter. 

The role of the filters is to allow the restriction of the sequences to which the objective is applied, thereby allowing us to guide over which part of the generation process the uncertainty is computed. While this could, in theory, be any subset of all possible sequences, we rather use algorithmic approaches in practice to determine the data we use for uncertainty calculations. Some of our filter functions define the filter criterion with reference to some other sequence of tokens $\mathbf{v}'\in \mathcal{V}$ as \textit{reference path}. We use the convention $f_{\mathbf{v}'}$ for such references, i.e., they are not a parameter of the filter but rather a property of the function. To get a better understanding of the filters, we now present examples for filters with a reference path (e.g., single token filter, prefix filter, and suffix filter) and without one (e.g., the EOS filter and the fixed-length filter).

\paragraph{End-of-output filter}
A simple filtering objective is that we are only interested in the evaluation of complete output sequences, without considering the interpretation. To achieve this, we can define the filter
\begin{equation}
f^{EOS_Y}(\mathbf{v})=
\begin{cases}
1 & \textit{if}~ \mathbf{v}~\textit{ends with}~EOS_Y\\
0 & \textit{otherwise}.
\end{cases}
\end{equation}

\paragraph{Single token filter}
A common perspective on uncertainty is not to look at a whole sequence, but rather only at the uncertainty of the generation of a single token. Formally, this means we have a path in the sampling tree of length $\mathbf{v}'=(v_1, ..., v_n)$ as reference and want to know the uncertainty associated with a token $v_n$. We can define a filter for this such that
\begin{equation}
f^{\textit{st}}_{\mathbf{v}'}(\mathbf{v}) = \begin{cases}
1 & \textit{if}~ \mathbf{v}_{1:|\mathbf{v}|-1}=\mathbf{v}'_{1:|\mathbf{v}'|-1} \wedge |\mathbf{v}|=|\mathbf{v}'|\\
0 & \textit{otherwise},
\end{cases}
\end{equation}
This filter only accepts sequences that have the same length as the reference, and where all but the last token must be equal. 

\paragraph{Fixed-length filter}
An extension of the single token filter that drops the requirement that the prefixes are the same is the fixed-length filter, defined as
\begin{equation}
f^{\textit{fl}}_n(\mathbf{v}) = \begin{cases}
1 & \textit{if}~ |\mathbf{v}|=n\\
0 & \textit{otherwise}.
\end{cases}
\end{equation}
With this filter, all sequences of a fixed length within the sampling tree can be selected. 

\paragraph{Prefix filter}
We can also generalize the above two filter functions into a generic filter that allows us to require a common prefix $\mathbf{v}' \in \mathcal{V}$ and then allow all possible suffixes as
\begin{equation}
f^{\textit{pre}}_{\mathbf{v}'}(\mathbf{v}) = \begin{cases}
1 & \textit{if}~ (v_1=v_1') \wedge \dots \wedge (v_{|\mathbf{v}'|}=v'_{|\mathbf{v}'|})\\
0 & \textit{otherwise}.
\end{cases}
\end{equation}

\paragraph{Suffix filter}
We can use the same principle, to define a suffix filter as
\begin{equation}
f^{\textit{suf}}_{\mathbf{v}'}(\mathbf{v}) = \begin{cases}
1 & \textit{if}~ (v_{|\mathbf{v}|-|\mathbf{v}'|+1}=v_{1}') \wedge \dots \wedge (v_{|\mathbf{v}|}=v'_{|\mathbf{v}'|})\\
0 & \textit{otherwise}.
\end{cases}
\end{equation}

\paragraph{Composite filter}
Filters can also be freely combined. Let $f^1, f^2$ be two filter functions. We then construct a composite filter $f^{cmp}$ as:
\begin{equation}
f^{\textit{cmp}}(\mathbf{v}) = f^1(\mathbf{v}) \cdot f^2(\mathbf{v}).
\end{equation}
This can be used to, e.g., combine a prefix filter with an end-of-output filter to only consider all possible outputs given a prompt. Since the composite filter is a filter as well, this can also be used to chain more than two filters together.

\subsection{Definition of objectives}

Uncertainty is considered with respect to a particular objective, i.e., a target concept with respect to which the uncertainty is considered. We enable this through a function that allows us to compute a score with respect to the relationship between a target sequence and another sequences. Formally, we consider an objective as a function
\begin{equation}
    o: \mathcal{V} \times \mathcal{V} \to \mathbb{R}^+_0
\end{equation}
i.e., from a pair of sequences over tokens to a non-negative real-valued number. By convention, the first of these two parameters is the target and the second can be any other sequences. Typically, these objectives are similarities mapping to $[0,1]$ (see examples in Section~\ref{sec:families-of-objectives}). However, they can also be general, positive scores. 

Using the objective function, together with a filter, we now define the \textit{objective-weighted probability} associated with any node $\mathbf{v}$ in the sampling tree as:
\begin{equation}
\label{eq:obj-prob}
    pr^{o,f}_{\mathbf{v}} = \frac{1}{\sum_{\mathbf{v}', \mathbf{v}'' \in \mathcal{V}^f} f(\mathbf{v}')o(\mathbf{v}', \mathbf{v}'') pr_{\mathbf{v}''}}\sum_{\mathbf{v'} \in \mathcal{V}^f} f(\mathbf{v})o(\mathbf{v}, \mathbf{v}') pr_{\mathbf{v}'}.
\end{equation}

The above effectively defines the objective as a relation between two arbitrary sequences of tokens and then uses this to compute the objective-weighted probability as weighted average of the probability over all sequences in the filter. The sum over $\mathbf{v}' \in \mathcal{V}^f$ evaluates the objective function between $\mathbf{v}$ and all $\mathbf{v}'$ that pass through the filter and uses this as weighting factor to create a weighted sum. The fraction is used as normalization by computing this weighted sum over all pairs of sequences to ensure the result is a probability distribution. This also implies the condition
\begin{equation}
\sum_{\mathbf{v}, \mathbf{v}' \in \mathcal{V}^f} f(\mathbf{v})o(\mathbf{v}, \mathbf{v}') pr_{\mathbf{v}} > 0
\end{equation}
on the combination of objective and filter function.

The concrete meaning of this abstract concept is now explained through concrete families of filters and objective functions, i.e., objective functions that share important properties.

\subsubsection{Families of objectives}
\label{sec:families-of-objectives}

While the objective can, in general, be any function that maps pairs of tokens to numbers, there are relatively few functions that have a meaningful application for the estimation of the uncertainty. Below, we consider three families of such functions as examples for suitable objectives: the syntactic equality, clusters of sequences, and the similarity between sequences.
For these families, we assume that we have binary filters as in our examples above, i.e., filter functions that map to $\{0,1\}$, to enable cleaner definitions. For non-binary filters, we effectively get the same results, but require additional multiplication with the filter function and normalization, which we avoid here for clarity.

\paragraph{Syntactic equality}

The simplest objective function is
\begin{equation}
o^{\textit{syn}}(\mathbf{v}, \mathbf{v}') = \mathds{1}[\mathbf{v}=\mathbf{v}']
\end{equation}

where $\mathds{1}$ is the characteristic function that is 1 if and only if the condition holds. Thus, this objective yields a one, if two sequences of tokens are equal. This effectively means that we have two strings with exactly the same syntax. With this objective function we get
\begin{equation}
pr_{\mathbf{v}}^{o^{\textit{syn}},f} = pr_{\mathbf{v}}
\end{equation}
for all $\mathbf{v}\in \mathcal{V}^f$. This is expected, because this is how LLMs actually sample outputs: the next syntactic token is selected without considering additional objectives like the similarity between tokens. 

\paragraph{Hard clustering}
A bit more general is that we may want to define objectives in which sequences form clusters, such that each sequence belongs to exactly one cluster, e.g., based on the semantic meaning. The clustering can be expressed by a function
\begin{equation}
\textit{hc}: \mathcal{V} \to C
\end{equation}
where $C=\{1, ..., |C|\}$ is a finite index set that represents the clusters. Using this function, we can define the objective function as
\begin{equation}
o^{\textit{hc}}(\mathbf{v}, \mathbf{v}') = \mathds{1}[\textit{hc}(\mathbf{v})=\textit{hc}(\mathbf{v}')].
\end{equation}

When we plug this objective function in the above definition, we get
\begin{equation}
pr_{\mathbf{v}}^{o^{\textit{hc}},f} = \frac{1}{|\{\mathbf{v}',\mathbf{v}'' \in \mathcal{V}^f: \textit{hc}(\mathbf{v}')=\textit{hc}(\mathbf{v}'')\}|} \sum_{\mathbf{v}' \in \mathcal{V}^f: \textit{hc}(\mathbf{v})=\textit{hc}(\mathbf{v}')}  pr_{\mathbf{v}'}.
\end{equation}

As can be seen, such objectives are suitable if we want to know the probability of sampling a sequence within a cluster.

\paragraph{Similarity / Soft clustering}

With texts, we are often also interested in how similar two texts are, e.g., operationalized with a cosine similarity between two embeddings. In general, this can be modeled through similarity functions
\begin{equation}
\textit{sim}: \mathcal{V} \times \mathcal{V} \to [0,1]
\end{equation}
which we can use to define an objective that takes the similarity between sequences into account as
\begin{equation}
o^{\textit{sim}}(\mathbf{v}, \mathbf{v}') = \textit{sim}(\mathbf{v}, \mathbf{v}'). 
\end{equation}
We then have
\begin{equation}
pr_{\mathbf{v}}^{o^{\textit{sim}},f} = \frac{1}{\sum_{\mathbf{v}',\mathbf{v}'' \in \mathcal{V}^f} \textit{sim}(\mathbf{v}',\mathbf{v}'')} \sum_{\mathbf{v}' \in \mathcal{V}^f} \textit{sim}(\mathbf{v}, \mathbf{v}') pr_{\mathbf{v}'}.
\end{equation}

In contrast to the hard clustering, all probabilities of the filtered tokens now affect the outcome, which is analogous to the weighted computation of the expected values for expectation maximization as soft clustering algorithm.

\subsection{Likelihood and information-theoretic concepts of uncertainty within the sampling tree}

There are two common ways to look at uncertainty. We can look at a specific outcome and compute the likelihood of this outcome, or we can look at the uncertainty of the decisions we make when computing an outcome. The former is the direct computation of the probability of the outcome, while the latter is covered by the information-theoretic concept of entropy. 

\subsubsection{Probabilities and objective-weighted probabilities}

For classification models, a common view on the uncertainty of the results is to look at the probability of a class. Since our sampling tree models and auto-regressive classification process, we can apply the same concept here. Using our sampling tree, this corresponds to the probability that is assigned to a single node. We already defined the probability that the token $v_i$ is generated given the tokens $\mathbf{v}_{1:i-1}$ as 
\begin{equation}
p(v_i|\mathbf{v}_{1:i-1}) = pr_{\mathbf{v}_{1:i}}.
\end{equation}

Consequently, we get
\begin{equation}
\label{eq:prob}
p(\mathbf{v}) = \prod_{i=1}^{|\mathbf{v}|} pr_{\mathbf{v}}
\end{equation}
for the probability of a sequence in our sampling tree and
\begin{equation}
p(\mathbf{v}_{k:n}|\mathbf{v}_{1:k-1}) = \prod_{i=k}^{|\mathbf{v}|} pr_{\mathbf{v}}
\end{equation}
for the probability of a $\mathbf{v}_{k:n}$ given known tokens $\mathbf{v}_{1:k-1}$. For example, we can use this to compute the uncertainty of the output $\mathbf{y}$ given a prompt $\mathbf{x}$ as
\begin{equation}
p(\mathbf{y}|\mathbf{x}) = \prod_{i=1}^{|\mathbf{y}|} pr_{\mathbf{x},\mathbf{y}_{1:i}}.
\end{equation}

Our definition so far ignored the filters and objectives. Because we defined both filters and objectives to be directly associated with nodes of the sampling tree, we can generalize the equations above using an objective function $o$ and a filter function $f$ by using the objective-weighted probability, e.g., 
\begin{equation}
\label{eq:prob-generic}
p^{o,f}(\mathbf{v}) = \prod_{i=1}^{|\mathbf{v}|} pr^{o,f}_{\mathbf{v}_{1:i}}.
\end{equation}
for the objective-weighted probability of the sequence $\mathbf{v}$.

With Equation \ref{eq:prob-generic} we have the means to compute the probability of any subsequence (e.g., a single token, the full output sequence given a prompt, or the uncertainty of the interpretation) with respect to any objective.\footnote{Note, that while we only discuss subsequences of consecutive tokens here, our concepts can also be directly applied to multiple non-consecutive subsequences: this would only require us to have one product per consecutive subsequence and multiply all these products with each other.}

\subsubsection{Entropy of sampling}

The consideration of probabilities has the weakness that it does not consider the probability of other outcomes. Since we already established in Section~\ref{sec:basic-notations} that we can interpret the sampling as random variables, we can use the concept of entropy from information theory to compute the uncertainty associated with the decision to select a token. Within our sampling tree, this means that we do not just look at a single node for a probability, but rather at all siblings. Formally, we can now define the entropy of the decision for selecting the $i$-th token of the output as the entropy of the random variable $V_i$ given already known tokens $\mathbf{v}_{1:i-1}$
\begin{equation}
H(V_i|\mathbf{V}_{1:i-1}=\mathbf{v}_{1:i-1}) = -\sum_{v \in V} pr_{\mathbf{v}_{1:i-1},v} \log pr_{\mathbf{v}_{1:i-1},v}.
\end{equation}

Same as above, we can extend this concept to the entropy of a sequence. This means that we need to consider the entropy of all decision made during the sampling process when generating a sequence. First, we consider this for the decisions that lead to a sequence $\mathbf{v}$. Since the auto-regressive process generates the outputs one step at a time, this means that for the selection of $v_i$ the decisions for $\mathbf{v}_{1:i-1}$ were already made, i.e., the entropy for $V_i$ is only based on the siblings of $v_i$. This means that we look at the entropy along the path $\mathbf{v}_{1:n}$ within the sampling tree, i.e., we consider all siblings of the nodes along this path. Formally, we define the entropy of the decisions $V_i$ for the generation of the sequence $\mathbf{v}$ as
\begin{equation}
\sum_{i=1}^{|\mathbf{v}|} H(V_i|\mathbf{V}_{1:i-1}=\mathbf{v}_{1:i-1})
= -\sum_{i=1}^{|\mathbf{v}|} \sum_{v \in V} pr_{\mathbf{v}_{1:i-1},v} \log pr_{\mathbf{v}_{1:i-1},v}
\end{equation}
The drawback of looking at specific sequences is that this ignores what could have happened when a different decision is made. For example, imagine a prompt $\mathbf{x}$ such that $P(Y_1=\text{``foo''}|\mathbf{X}=\mathbf{x})=0.49$ and $P(X_1=\text{``bar''}|\mathbf{X}=\mathbf{x})=0.49$. When we generate ten sequences for this prompt, about half of them start with ``foo'', and half with ``bar''. The entropy for each of the specific sequences would only consider the subtree of the sampling tree for ``foo''  or ``bar'',  respectively. This would not be a realistic estimation of the entropy of the generation process given the prompt, as this would need to consider both subtrees within the sampling tree. Thus, when we want to model the uncertainty of the whole sequence generation process, that considers all possible decisions, we actually have to base this on the whole sampling tree, as any node can be selected with a certain probability. Formally, we define the entropy of the sequence $\mathbf{V}_{k:n}$ given known tokens $\mathbf{v}_{1:k-1}$ as
\begin{align}
H(\mathbf{V_{k:n}}|\mathbf{V}_{1:k-1}=\mathbf{v}_{1:k-1}) 
&=\sum_{i=k}^n H(V_i|\mathbf{V}_{1:k-1}=\mathbf{v}_{1:k-1},\mathbf{V}_{k:i})\\
&=-\sum_{\mathbf{v}' \in V^n: \mathbf{v}'_{1-k}=\mathbf{v}_{1-k}} 
\left( \prod_{i=k}^n pr_{\mathbf{v}'_{1:i}} \log \prod_{i=k}^n pr_{\mathbf{v}'_{1:i}} \right).
\end{align}

For example, with $\mathbf{V}_{1:k-1}=\mathbf{X}$ and $\mathbf{V}_{k:n}=\mathbf{Y}$, we get the entropy of the output sequence $\mathbf{Y}$ given a prompt $\mathbf{X}$ as
\begin{align}
H(\mathbf{Y}|\mathbf{X}=\mathbf{x}) 
&=\sum_{i=1}^{n_y} H(Y_i|\mathbf{X}=\mathbf{x}, \mathbf{Y}_{1:i-1})\\
&=-\sum_{\mathbf{y}\in V^{n_y}} \left( \prod_{i=1}^{n_y} pr_{\mathbf{x},\mathbf{y}_{1:i-1}} \log \prod_{i=1}^{n_y} pr_{\mathbf{x},\mathbf{y}_{1:i-1}} \right).
\end{align}

All of these concepts can be generalized to \textit{objective-weighted entropy} by simply using the objective-weighted probability for the computation for the entropy, e.g., as
\begin{align}
H^{o,f}(\mathbf{V_{k:n}}|\mathbf{V}_{1:k-1}=\mathbf{v}_{1:k-1}) 
&=\sum_{i=k}^n H^{o,f}(V_i|\mathbf{V}_{1:k-1}=\mathbf{v}_{1:k-1},\mathbf{V}_{k:i})\\
&=-\sum_{\mathbf{v}' \in V^n: \mathbf{v}'_{1-k}=\mathbf{v}_{1-k}} 
\left( \prod_{i=k}^n pr^{o,f}_{\mathbf{v}'_{1:i-1}} \log \prod_{i=k}^n pr^{o,f}_{\mathbf{v}'_{1:i-1}} \right).
\end{align}

\subsection{Estimators}
\label{sec:estimators}

Since the vocabulary size used for LLMs is typically very large (e.g., OpenAI uses tiktoken\footnote{\url{https://github.com/openai/tiktoken}} with 199,997 tokens), the exact computation of the uncertainty of the sequence generation processes may prohibitively expensive, as in extreme cases we would have to consider the whole sampling tree. Estimators are used to avoid this by restricting which parts of the sampling tree we consider. We can extend our framework with a second filter $\hat{f}:\mathcal{V} \to \mathbb{N}_0$ and compute an \textit{estimated objective-weighted probability} through a small modification of Equation \ref{eq:obj-prob} as

\begin{equation}
p^{o,f, \hat{f}}_{\mathbf{v}} = \frac{1}{\sum_{\mathbf{v}', \mathbf{v}'' \in \mathcal{V}^{f, \hat{f}}} f(\mathbf{v}')\hat{f}(\mathbf{v}')o(\mathbf{v}, \mathbf{v}') pr_{\mathbf{v}'}} \sum_{\mathbf{v'} \in \mathcal{V}^{f, \hat{f}}} f(\mathbf{v})\hat{f}(\mathbf{v})o(\mathbf{v}, \mathbf{v}') pr_{\mathbf{v}'}.
\end{equation}

with $\mathcal{V}^{f, \hat{f}} = \{\mathbf{v} \in \mathcal{V}: f(\mathbf{v})>0 \wedge \hat{f}(\mathbf{v})>0\}$. 
With different filters, we can then define different ways to estimate objectives. 

\subsubsection{Top-\texorpdfstring{$k$}{k} tokens}

A common approach is to restrict the branching factor of the sampling tree~\citep{fan-etal-2018-hierarchical}. Instead of always considering all tokens, we can rather consider only the Top-$k$ tokens that have the largest probability. Formally, we define
\begin{equation}
\hat{f}^{topk}(\mathbf{v}) = \begin{cases}
1 & if~ \bigwedge_{i \in 1, ..., n} v_i\in \arg\max^k_{v' \in V} pr_{v_{1:{i-1}},v'}\\
0 & \textit{otherwise}
\end{cases}
\end{equation}
with $\mathbf{v}=(v_1, ..., v_n)$. This effectively reduces the branching factor of the tree to $k$, resulting in a filtered sampling tree size of $k^n$ for sequences of length $n$. While still exponential for $k>1$, we now have much better control over the actual size of the tree, especially when used for estimating objectives that already do not consider the whole tree, e.g., single token-filters, prefix, or suffix filters. 

\subsubsection{Probability mass filter}

The drawback of the Top-$k$ filter is that we do not know how much probability mass is actually captured by the estimator, which means it is hard to reason about the reliability of the estimation. In an extreme case, i.e., the token probabilities are uniform, we would only capture $\frac{k}{|V|}$ of the probability mass. On the other hand, if the distribution is heavily skewed, even a single token could be sufficient to capture 99\% of the probability mass. To mitigate this, we can define a dynamic filter that is defined such that it always takes the most likely tokens until at least $pm\in [0,1]$ of the probability mass are taken into account in the manner of nucleus sampling~\citep{holtzman2020curious}. To define such a filter, we first define
\begin{equation}
V^{pm}(\mathbf{v})=\left\{ V' \subseteq V: \left(\sum_{v \in V'} pr_{\mathbf{v},v}>pm\right)~\wedge \left(\forall v'\in V' ~\forall v''\in V\setminus V':pr_{\mathbf{v},v'}\leq pr_{\mathbf{v,}v''}\right)\right\}
\end{equation}
as all possible sets of children of the sequence $\mathbf{v}$ in the sampling tree that have at least probability mass $pm$. The second condition ensures that these sets include the most likely tokens. We can identify the required branching factor using the smallest such set, i.e.,
\begin{equation}
 k^{pm}_n = min_{V \in V^{pm}(\mathbf{v}_{1:n-1})} |V|   
\end{equation}
We can then select a smallest set of tokens with the required probability mass\footnote{Assuming ties are possible, i.e., multiple tokens with the same probability, there is a possibility that there are multiple subsets of tokens that have the minimal size, i.e., containing different of these tokens with a tie. This is only the case, if these are also the tokens with the smallest probability in the subset. In this case, a random subset would need to be picked. For our notation, this is not relevant, since we simply use the element-operator instead.} as
\begin{equation}
V_n^{pm}(\mathbf{v}_{1:n-1}) \in \left\{V \in V^{pm}(\mathbf{v}_{1:n-1}): |V|=k_n^{pm}\right\}   
\end{equation}
Finally, we get the filter as
\begin{equation}
\hat{f}^{pm}(\mathbf{v}) = \begin{cases}
1 & if~\bigwedge_{i \in 1, ..., n} v_i \in V_n^{pm}(\mathbf{v}_{1:n-1})\\
0 & otherwise.
\end{cases}
\end{equation}

\subsubsection{Monte Carlo simulation}
\label{sec:monte-carlo}

With Monte Carlo simulation, we simply sample multiple times. When doing so, we can, e.g., keep the prompt fixed to study the uncertainty of the output, but also use different prompts, e.g., to study prompt sensitivity by using different filters. The Monte Carlo simulation is then restricted to the unfixed parts of the sampling tree. Formally, the sampling tree is a countable structure, which implies that there is an enumeration $\textit{enum}: \mathcal{V} \to \mathbb{N}$ of all possible sequences. We can then express $r\in \mathbb{N}$ samples generated with Monte Carlo simulation through a map
\begin{equation}
    \textit{mc}: \mathcal{V} \to \mathbb{N}_0 \quad s.t. \quad \sum_{i=1}^\infty \textit{mc}(i)=r
\end{equation}
and define a filter
\begin{equation}
\hat{f}^{\textit{mc}}(\mathbf{v}) = \textit{mc}(\textit{enum}(\mathbf{v})).
\end{equation}

\section{Operationalization}
\label{sec:operationalization}
To demonstrate that our framework is viable and gives us a well-defined common ground for uncertainty quantification of LLMs, we show how papers from the literature can be expressed by choosing appropriate objectives and filters, as well as limitations of our work.  

\subsection{Existing approaches in the related work}

We now describe how to use filters and objective to describe different types of methods from the state-of-the-art. We sometimes describe that multiple filters can be used. In this case, we assume without loss of generality that they are combined using a composite filter. Further, we will always refer to the objective-weighted probability and objective-weighted entropy below, when speaking about expressing probabilities and entropies within our framework. 

The first set of methods from the literature focuses on the uncertainty of the generated output $\mathbf{Y}$, without any consideration of the interpretation $\mathbf{Z}$. To ignore the interpretation, we can use a end-of-output filter $f^{EOS_Y}$. To make the analysis apply to a specific prompt -- including the consideration of the uncertainty when using in-context learning strategies like \cite{ling-etal-2024-uncertainty} --, a prefix filter $f_{\mathbf{X}}^{pre}$ can be used. To capture a basic notion of uncertainty by the probability of the last token~\citep{jiang2021}, we add a single token filter $f_{\mathbf{v}'}^{st}$. Similarly, we can model the entropy of generated output sequences~\citep[e.g.,][]{kadavath2022language, balabanov2025uncertainty} within our framework in the same way, we just need to drop the single token filter. All of the aforementioned approaches look at exact equality, i.e., we use the syntactic equality $o^{syn}$ as objective.

The notation of semantic entropy as introduced by \cite{kuhn2023semantic} and \cite{farquhar2024detecting} can also be expressed within our framework. Here, the idea is to group semantically equal outcomes together, which is exactly what the hard-clustering objective functions achieve. Thus, these techniques naturally map into our framework. We can also further extend our framework to uncertainty of outcomes given a specific prompt. The kernel language entropy by \cite{nikitin2024kernel} and other similarity-based variants of the entropy \citep[e.g.,][]{duan-etal-2024-shifting} can instead be expressed as a soft-clustering objective, where the kernel that describes the semantic similarity between the sequences becomes the objective. 

All of the above mentioned approaches that consider the entropy, use a Monte Carlo simulation as described in Section \ref{sec:monte-carlo} for the estimation of entropy, i.e. effectively yield the estimated objective-weighted entropy using a $\hat{f}^{mc}$ filter. An estimation method as proposed by \cite{aichberger2024rethinkinguncertaintyestimationnatural} can be implemented by a Top-$k$ estimation filter with $k=1$. 

Self-consistency is another perspective that is often considered for uncertainty~\citep[e.g.,][]{arabzadeh2025human, huang2025look}. Here, the focus is rather on a set of prompts $\mathcal{X} \subset \mathcal{V}$  and the output $\mathbf{Y}$. We can define a simple variant of the prefix filter to restrict the sampling tree to this family of prompts:
\begin{equation}
f_\mathcal{X}(\mathbf{v})=
\begin{cases}
1 & if~ \sum_{\mathbf{v}\in \mathcal{X}} f^{pre}_{\mathbf{v}}(\mathbf{v})>0 \\
0 & otherwise
\end{cases}
\end{equation}
Combined with the end-of-output filter $f^{EOS_Y}$, this means that we only consider the outputs for a given set of prompts. Using the syntactic equality $o^{syn}$, we can then, e.g., compute the probability that an LLM outputs a certain category as input for Cohen's $\kappa$ as required by \cite{arabzadeh2025human}. Alternatively, the similarity of outputs can be assessed using a semantic measure -- same as above -- by a defining a soft-clustering objective function based on the similarity measures, e.g., for measures using BLEU as by~\cite{huang2025look}.

The consideration of the uncertainty of the interpretation $\mathbf{Z}$ is typically not considered from a formal perspective that looks that probabilities and the entropy of judgments~\citep{gu2024survey} -- though there are first works working towards a more formal consideration of this~\citep{da2025understanding} -- but rather by reconstructing reasoning within explanations instead of directly measuring the uncertainty. However, in general the focus seems to be fully on quality of outputs through confusion-matrix analysis \citep[e.g.,][]{wagner2024black}. Technically, we can express such analysis within our framework: the confusion matrix is an estimator for class probabilities and we could model the classes through a hard-clustering objective function. 

For the methods above, we often did not distinguish between whether the uncertainty measurement is restricted to a single token or a sequence of tokens. However, with our framework, this distinction is trivial: for full sequences, an EOS filter is used, when specific lengths are desired, we can use more restrictive filters like single token filters, fixed-length filter or any other method we want to prune the output space with. 

\subsection{Beyond the probability and entropy: limitations of our framework}

Not all works in the literature directly use the probability or the entropy, but rather, e.g., use the negative log-likelihood~\citep{manakul-etal-2023-selfcheckgpt}, do more computations with the token probability or output entropy, to scale the outputs in a certain manner~\citep{fadeeva-etal-2024-fact}, or use a Bayesian view on uncertainty~\citep{wang2025subjective}. The estimation of the underlying probabilities can be done within our framework through the objective-weighted probability and entropy and the framework could easily be extended with an, e.g., objective-weighted negative log likelihood. 

We do not cover verbalizing methods~\citep[e.g.,][]{lin2022teaching} were LLMs report uncertainty directly within the output sequence $\mathbf{Y}$ with our framework. However, we believe this is a reasonable limitation, especially given the tendency of LLMs to be overconfident in their results which is reinforced through the instruction-tuning process~\citep{kadavath2022language}.

\section{Discussion}
Next, we discuss potential implications of our framework for future work before specifically reflecting on possible extensions to model and measurement uncertainties.

\subsection{Implications of our framework for future work}

The advantage of a unified formal framework is not only that we can better see how different methods are related with each other, but also that we can see which aspects might be worth future consideration. Keeping in line with our work, we discuss this based on the objective and filter functions. 

The obvious gap that our analysis highlights is the lack of the consideration of the interpretation $\mathbf{Z}$ through a formal uncertainty lens that takes the likelihood and entropy into account. This is both relevant from a human studies perspective (do humans mitigate or amplify LLM uncertainty or is this unchanged), but also for agentic AI approaches, where the interpretation of outputs from other LLMs is crucial, raising the question if uncertainty is mitigated, propagated unchanged, or amplified. Ideally, we would go towards a holistic analysis of the uncertainty that considers the whole sampling tree to understand the uncertainty implications in agentic setups. 

Beyond this clear gap, we also observe that the common ground that our framework yields with respect to the underlying mathematical principles of the uncertainty estimation helps us see similarities between approaches. Importantly, we show that the distinction between the semantic entropy methods is actually only a difference in the types of clustering used~(see Section~\ref{sec:families-of-objectives}). This can be used to understand when which kind of method is preferable: 
\begin{itemize}
    \item Tasks that have clearly defined and well-separated categories (e.g., multiple-choice questions) should use hard-clustering objectives when studying the uncertainty.
    \item Tasks that lack clear ground-truth and have a more gradual scale for result quality (e.g., essay writing, summarization) should use soft-clustering objectives.
\end{itemize}

When looking at the filters, we observe that methods that study uncertainty with respect to the prompting use a variant of a prefix-filter that restricts the prompts to certain variations. On the other hand, methods that study how likely a certain outcome is (e.g., for the estimation of Cohen's $\kappa$) rather rely on a suffix filter. We also already saw that both can be combined, e.g., to see uncertainty of fixed outcomes with respect to prompt variations~\citep{arabzadeh2025human}. However, current taxonomies \citep[e.g.,][]{xia2025survey} largely ignore this relationship between methods from a structural perspective. A systematic taxonomy that maps the formal view (filtering with specific families) to the uncertainty aspect that is studied (e.g., prompt sensitivity focusing on $\mathbf{X}$, uncertainty of output focusing on $\mathbf{Y}$, uncertainty on interpretations focusing on $\mathbf{Z}$) is still missing. Our work establishes the formal foundations required for this. 

Taking both aspects together, an overall taxonomy that defines 
\begin{itemize}
\item which type of objective function (syntactic, hard-clustering semantic similarity, soft-clustering semantic similarity) is suitable for which underlying tasks, as well as 
\item which filters are required to study which aspects of uncertainty, 
\end{itemize}
would result in a toolbox for the systematic consideration of uncertainty with suitable methods for all contexts. This tool-box would automatically encompass the current state-of-the-art, but additionally allow for new variants including the consideration of the uncertainty of interpretations. 

Another aspect that is not yet considered systematically in the state-of-the-art is the impact on the estimation method for the entropy on the outcomes. As we describe above, Monte Carlo sampling is the de facto standard for entropy estimations. However, as we show there are other approaches as well with different advantages, because they can, e.g., give us guarantees about the probability mass that is considered (probability mass filter). We also do not know to which degree beam search methods (as an variation of the Top-$k$ estimator) could help with better estimations of uncertainty by estimating the uncertainty among a set of highly-likely sampling paths. Notably, these additional considerations are also practically relevant to understand the uncertainty associated with non-greedy decoding strategies~\citep{aichberger2024rethinkinguncertaintyestimationnatural}. 

Moreover, since we show that estimation methods are nothing more than filters, the interaction between the estimation method and other, task-oriented filters needs to be considered. The interactions include runtime aspects (e.g., more permissive filters for tasks require more inferences, increasing the runtime) but also establishing the expected quality of estimation filters (e.g., more permissive filters that consider multiple prompts will still require multiple inferences per prompt to study the decoding-uncertainty with Monte Carlo simulation). 

\subsection{Model and measurement uncertainty}
\label{sec:measurement-uncertainty}

All of the above, assumes that we have a fixed set of parameters determining our (estimated) objective-weighted probabilities, i.e., fixed models $\Theta_X, \Theta_Y, \Theta_Z$ that define the probabilities for $\mathbf{X}, \mathbf{Y}$, and $\mathbf{Z}$ in the sampling tree, a filter $f$, a single objective $o$, and possibly an estimation filter $\hat{f}$. However, there is also uncertainty associated with these parameters:
\begin{itemize}
    \item The model $\Theta_Y$ depends on the architecture choice, model size, training parameters, and training data, as well possible decoding parameters. These can be influenced by human decision-making (e.g., ``we use the commonly used ...'') or through algorithmic processes (e.g. auto-ML approaches).
    \item The models $\Theta_X$ and $\Theta_Z$ could either model the probability of human actions or also be LLMs. For LLMs, the same considerations as for $\Theta_Y$ apply. With human actions, there is also further uncertainty, e.g., due to differences in educational background. 
    \item The filters (incl. the estimation filter) and objectives determine and constitute the measurement uncertainty. As already indicated above, the choice of estimation might impact the result. Similarly, a different measure for the semantic similarity as objective (e.g., cosine similarity with SentenceBERT vs. BLEU) could impact the results. 
\end{itemize}

An example that takes the model uncertainty on the decoding into account is the work by \cite{balabanov2025uncertainty}: they parametrized the model $\Theta$ by the training to assess how different training mechanisms affect the uncertainty. While we did not directly include this within our framework, as this goes beyond the scope of our work, this can be seen as an orthogonal extension. Our framework can be naturally extended to include the additional uncertainties, when we consider them as random variables as well. We could then, condition on these variables. For example, to account for model uncertainty, we can change Equation \ref{eq:path-prob} that defines the probabilities of paths to
\begin{equation}
pr_{\mathbf{v}_{1:i}} = p(v_i|\mathbf{v}_{1:i-1}, \Theta_X, \Theta_Y, \Theta_Z)    
\end{equation}
getting a more general definition of the sampling tree of which the version with fixed models from this paper is a special case. With a similar change to the objective-weighted probability, we could take the measurement uncertainty into account. 

Additionally, the assumption that the whole process is auto-regressive can be considered as a general uncertainty that our measurement construct introduces. Sampling processes that are not left-to-right or human judgment that may produce a whole sequence at once cannot be directly expressed with our current formalism. However, if we were to modify the sampling tree such that the nodes could also be sequences of tokens, such that the probability of the node would then be the joint probability of the whole sequence, conditioned on the ancestors of the tree, we could solve this problem. Alternatively, one could introduce ordering functions such that the order of the nodes in the tree is not necessarily equal to the order of the nodes that are generated. Both aspects would also be pure natural extensions of our work that lead to a more complex underlying model, where the current framework would be a special case with only single tokens as nodes, respectively with the tree in the same order as the sequence of tokens. 

\section{Conclusion}

In this work, we introduced a formal framework for analyzing uncertainty in large language model (LLM) text generation that unifies prompting, decoding, and interpretation. By modeling these stages as components of a sampling tree and introducing filter and objective functions, we provide a flexible and expressive foundation for representing a wide range of existing uncertainty quantification methods. This perspective not only reveals common underlying principles, such as the role of different types of clustering for variants of the semantic uncertainty, but also clarifies how different approaches relate to each other in a principled manner.

Our framework highlights important gaps in the current literature, particularly the limited treatment of interpretation uncertainty and the lack of systematic consideration of estimation methods and their interactions with uncertainty measures. By framing these aspects within the same formalism, we show opportunities for a more comprehensive analyses of LLM behavior. Future work can build on this foundation to develop standardized taxonomies, explore the yet-unexplored aspects we determined, and possibly extend this to complex, multi-agent settings.

\section*{Acknowledgement}
We thank Jonathan Drechsel and Max Klabunde for their helpful feedback on this work.

\bibliographystyle{tmlr}
\bibliography{bib}

\end{document}